\documentclass{article} %
\usepackage{colm2024_conference}

\usepackage{booktabs}
\usepackage{graphicx}
\usepackage{enumitem}
\usepackage{wrapfig}
\usepackage{algorithm}
\usepackage{algpseudocode}
\usepackage{wrapfig}
\usepackage{float}
\usepackage{microtype}
\usepackage{amsmath}
\usepackage{amssymb}
\usepackage{colortbl}
\usepackage[utf8]{inputenc}
\definecolor{lightgray}{rgb}{0.9,0.9,0.9}
\usepackage{caption}
\usepackage{subcaption}
\usepackage{xcolor}
\usepackage{setspace}
\usepackage{url}
\usepackage{multirow}
\usepackage{colortbl}
\usepackage{tabularx}
\usepackage{blindtext}
\usepackage{pgfplots}
\pgfplotsset{compat=1.18} 
\usepackage{tikz}
\usetikzlibrary{er,positioning,bayesnet}
\usepackage{makecell}
\usepackage{tipa}
\usepackage{siunitx}
\usepackage{nicefrac}
\usepackage{tocloft}
\usepackage{listings}
\usepackage[raster,skins]{tcolorbox} %
\usepackage{xltabular}
\usepackage{adjustbox}
\usepackage{xurl}
\usepackage{multicol}

\usepackage{amsmath,amsfonts,bm}

\def\eqref#1{equation~\ref{#1}}

\def\1{\bm{1}}

\DeclareMathAlphabet{\mathsfit}{\encodingdefault}{\sfdefault}{m}{sl}
\SetMathAlphabet{\mathsfit}{bold}{\encodingdefault}{\sfdefault}{bx}{n}

\makeatletter
\DeclareRobustCommand\onedot{\futurelet\@let@token\@onedot}
\def\@onedot{\ifx\@let@token.\else.\null\fi\xspace}

\def\eg{\emph{e.g}\onedot}

\makeatother

\newcommand{\shortname}{Hunyuan3D-Omni\xspace}

\title{\shortname: A Unified Framework for Controllable Generation of 3D Assets}

\author{
\bf Tencent Hunyuan3D 
}

\begin{document}

\maketitle

\begin{figure}[h]
\centering
\includegraphics[width=0.99\linewidth]{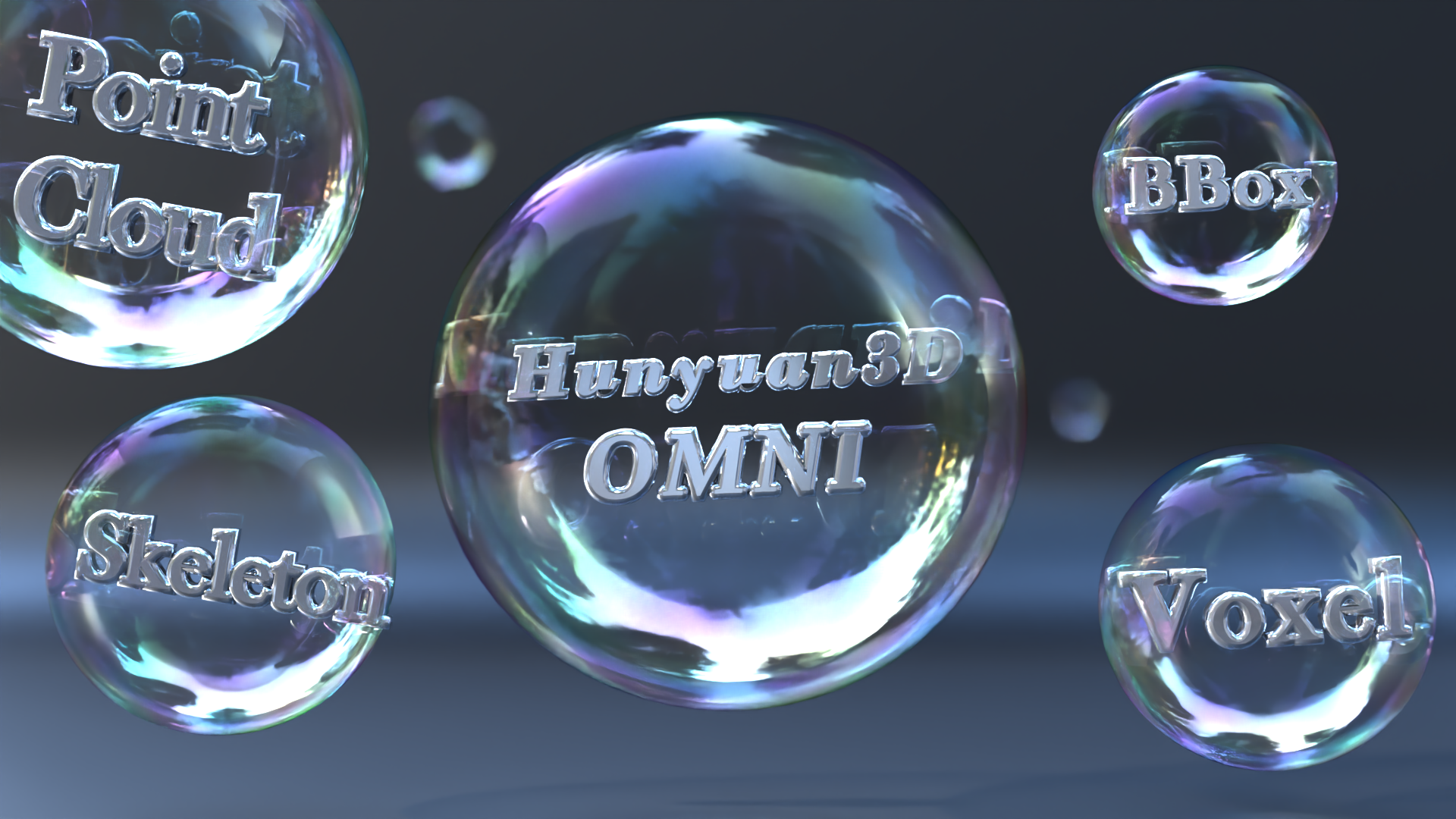}
\caption{\shortname is a unified framework for supporting controllable generation based on point cloud, bounding box, voxel, and skeleton.}
\label{fig:teaser-top}
\end{figure}

\begin{abstract}

Recent advances in 3D-native generative models have accelerated asset creation for games, film, and design. However, most methods still rely primarily on image or text conditioning and lack fine-grained, cross-modal controls, which limits controllability and practical adoption. To address this gap, we present \shortname, a unified framework for fine-grained, controllable 3D asset generation built on Hunyuan3D 2.1. In addition to images, \shortname accepts point clouds, voxels, bounding boxes, and skeletal pose priors as conditioning signals, enabling precise control over geometry, topology, and pose. Instead of separate heads for each modality, our model unifies all signals in a single cross-modal architecture. We train with a progressive, difficulty-aware sampling strategy that selects one control modality per example and biases sampling toward harder signals (~\eg, skeletal pose) while downweighting easier ones (~\eg, point clouds), encouraging robust multi-modal fusion and graceful handling of missing inputs. Experiments show that these additional controls improve generation accuracy, enable geometry-aware transformations, and increase robustness for production workflows.
\end{abstract}

\section{Introduction}
3D generation is a fundamental task in computer vision and computational imaging, with applications spanning virtual reality, gaming, and film. As the volume of 3D data continues to grow, native 3D generation has emerged as a mainstream approach, offering significant advantages in both quality and speed. It is anticipated that, with the ongoing expansion of high-quality 3D datasets, 3D generation models will evolve into the next generation of automated modeling tools, facilitating faster workflows and dynamic interactions in digital content creation.

Native 3D generation primarily involves two key components: the 3D Variational Autoencoder (VAE) and the 3D latent diffusion model (LDM). For example, a method utilizing VecSet representation employs a 3D VAE to compress point clouds into a VecSet, from which a decoder retrieves the Signed Distance Function (SDF) field for the 3D model. An iso-surface sampling technique is then applied to generate the visible 3D model from the SDF field. Similarly, the 3D LDM builds on the VecSet representation by stacking multiple layers of Diffusion Transformers (DiT) to facilitate the learning process from images to their corresponding 3D representations in VecSet form. Recent advancements, such as Hunyuan3D 2.1~\cite{hunyuan3d2025hunyuan3d21imageshighfidelity}, have showcased the powerful capabilities of native 3D generation for efficient and high-quality 3D modeling~\cite{zhao2024michelangelo,li2024craftsman}.

Despite these significant advancements, generating 3D assets from a single image remains an ill-posed problem, complicating the accurate reconstruction of complete 3D structures. This often results in uncertainties and ambiguities in 3D geometry generation. To enhance geometric accuracy, it is crucial to incorporate additional information through controllable generation techniques. Such techniques not only improve geometric fidelity but also enable customized outputs by imposing specific conditions. For instance, the integration of depth information can alleviate geometric distortions and spatial misalignments caused by viewpoint variations and self-occlusion, while also enriching geometric details. Recent studies, including Clay~\cite{zhang2024clay} and PoseMaster~\cite{yan2025posemaster}, have made notable progress in introducing additional conditions for downstream editing and pose control within the realm of 3D native generation models. Nevertheless, there remains a need for further exploration in developing a systematic and unified 3D controllable model.

In this paper, we introduce \shortname, a unified framework for fine-grained and controllable 3D asset generation. Building upon the foundational model Hunyuan3D 2.1 and following the workflow of 2D controllable generation models, \shortname enhances controllability and geometric accuracy by integrating various additional conditions, including point clouds, voxels, bounding boxes, and skeletons. To optimize training and model deployment costs, we consolidate these additional conditions into a single generative model. Specifically, we utilize point clouds to represent these extra conditions and propose a unified control encoder to differentiate between them and obtain corresponding embeddings. To preserve the structure and functionality of the base model, we concatenate the extracted embeddings with the DINO features of the input image. This approach allows us to achieve controllable 3D generation with minimal training steps. Experimental results demonstrate that \shortname effectively addresses common challenges in native 3D generation, such as distortions, flatness, missing details, and aspect ratio discrepancies, by providing additional control signals. Furthermore, it facilitates the standardization of character poses and the stylization of generated outputs, offering new perspectives and solutions for post-training applications in 3D generation.

\section{Related Work}

\subsection{3D Native Generation}
In recent years, the field of 3D generation has advanced rapidly, bringing significant impact to domains such as gaming, film, and animation. Early works include SDS and multi-view supervision approaches ~\cite{poole2022dreamfusion, liu2023zero, liu2023one, liu2024one, voleti2025sv3d} leverage pretrained image diffusion models to supervise the optimization of radiance fields~\cite{mildenhall2021nerf} or NeuS~\cite{wang2021neus}, thereby enabling 3D object generation. However, these methods suffer from multi-view consistency issues and slow generation speed, often producing noisy geometry. To address efficiency, LRM~\cite{hong2023lrm} introduces a feed-forward architecture that directly outputs a tri-plane radiance field, enabling fast single-image-to-3D generation. Subsequent works ~\cite{yang2024hunyuan3d, tang2024lgm, xu2024instantmesh, zhang2024gs} adopt similar strategies, but their results remain limited in geometric detail and texture fidelity.

3DShape2VecSet~\cite{zhang20233dshape2vecset} introduces a point cloud–based VAE to build a native 3D generation framework combining VAE and LDM, offering higher quality and faster generation. Michelangelo~\cite{zhao2024michelangelo} further aligns the VecSet latent space with semantics to enable text-conditioned generation. Many later works follow the VecSet paradigm, scaling up models and datasets, replacing EDM with flow matching, and incorporating MoE architectures. These techniques have yielded high-quality 3D generation models such as CLAY~\cite{zhang2024clay}, Craftsman3D~\cite{li2024craftsman}, TripoSG~\cite{li2025triposg}, FlashVDM~\cite{lai2025flashvdm}, and Hunyuan3D 2.0/2.1/2.5~\cite{zhao2025hunyuan3d, hunyuan3d2025hunyuan3d21imageshighfidelity, lai2025hunyuan3d2_5}, which can rapidly generate high-quality 3D assets from modalities such as single images or text. Another line of research utilizes voxel-based VAE representations to achieve similarly high-quality 3D generation, including works such as XCube~\cite{ren2024xcube}, TRELLIS~\cite{xiang2024structured}, and SparseFlex~\cite{he2025sparseflex}. Meanwhile, some works build upon these models, enabling low-poly generation~\cite{weng2024scaling}, part generation~\cite{yan2025x, zhang2025bang}, scene generation~\cite{yao2025cast}, material generation~\cite{jiang2025flexitex, he2025materialmvp, feng2025romantex}, and assets generation pipeline~\cite{lei2025hunyuan3d}, supporting many downstream applications. 

Although these methods enable fast generation of 3D objects from images or text, they generally lack support for more complex conditioning signals, such as points, bounding boxes, or voxels, which limits controllability and practical adoption.
\subsection{3D Controllable Generation}
Recently, 2D controllable generation has made remarkable progress in image synthesis and editing. Methods such as ControlNet~\cite{zhang2023adding}, T2I-Adapter~\cite{mou2024t2i}, and IP-Adapter~\cite{ye2023ip} introduce various structured conditions (e.g., edges, depth, and pose) into diffusion models, enabling high-precision control over generation results and highlighting the great potential of multimodal conditions in enhancing controllability and practicality.

In the area of 3D generation, similar approaches have emerged to support multiple types of control inputs. CLAY~\cite{zhang2024clay} explores the integration of point cloud, bounding box, and voxel conditions into generative models via LoRA finetuning, thereby adapting to different input modalities. PoseMaster~\cite{yan2025posemaster} incorporates pose control signals, enabling precise control over the pose of generated 3D characters. These works collectively demonstrate that accurate controllability is also achievable in 3D generation.

However, most existing research focuses on single or limited conditions, and a unified multi-condition framework for 3D controllable generation is still lacking. How to flexibly integrate diverse control signals such as points, voxels, bounding boxes, and poses within a single model to achieve fine-grained and cross-modal controllability of 3D contents remains an important yet insufficiently explored challenge.

\section{Method}
\subsection{Preliminaries}
{\bf 3D VAE}. Given an input point cloud $P \in R^{N \times (3 + C)}$ sampled from the mesh surface, where $C$ denotes surface normals, 3D VAE first extract point features and then obtain the corresponding latent vector set $Z \in R^{L \times d}$ via resampling from estimated distribution, where $L$ and $d$ indicate the length and dimension of latent VecSet, respectively. Subsequently, a decoder is applied to reconstruct the signed distance function (SDF) field $F_{sdf}$, in which we can leverage the iso-surface extraction to obtain explicit mesh output. The procedure of VAE can be formulated as follows:
\begin{align}
Z = \mathcal{E}(P), F_{sdf} = \mathcal{D}(Z)
% \end{align}
\end{align}

{\bf 3D Diffusion}. Given an image and its latent set representation $Z$ of a shape, the 3D diffusion model aims to model the denoising process, thereby achieving conditional generation from an arbitrary image. It first leverages an image encoder, such as DINO-v2~\cite{}, to capture image embeddings $c_i$ and then exploits the multi layers of DiT to predict the added noise or velocity. For a flow matching model used in Hunyuan3D 2.1~\cite{}, its training objective is to transform a simple noise distribution $x_0 \sim \mathcal{N}(0, I)$ into a complex data distribution $x_1 \sim D$ conditioned on image embeddings $c_i$, which can be formulated as follows:
\begin{equation}
\mathbb{E}_{t, {x}_0, {x_1}, c}\vert\vert{v}_\theta({x}, t, c)-(x_1-x_0)\vert\vert_2^2
\end{equation}

\begin{figure}[h]
\centering
\includegraphics[width=0.99\linewidth]{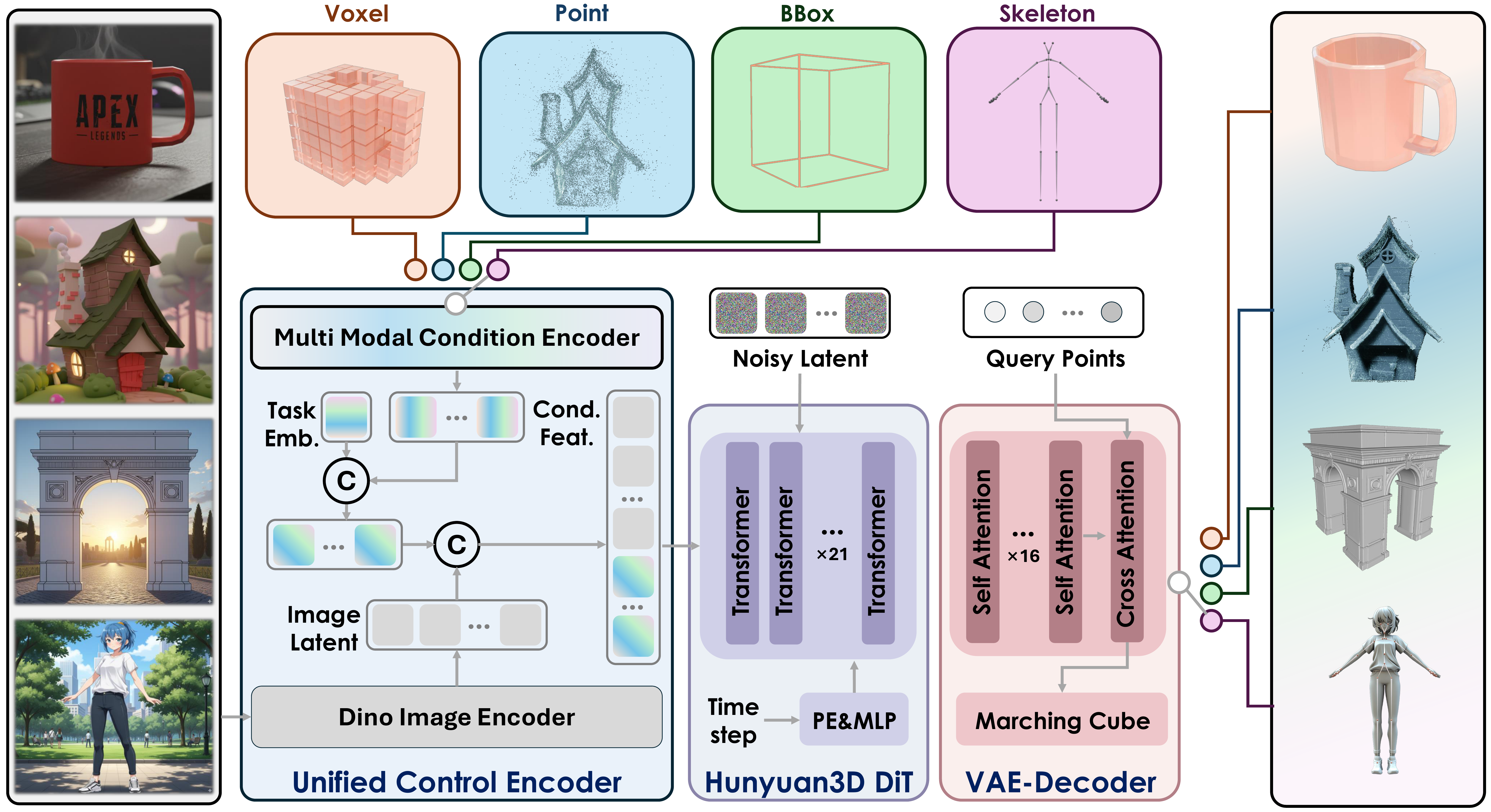}
\caption{Overview of \shortname Architecture. This framework integrates four conditioning modalities—point clouds, bounding boxes, voxels, and skeletons—through a unified control encoder. \shortname leverages a Diffusion Transformer (DiT) and VAE-based decoder, following the Hunyuan3D 2.1 framework, to generate high-quality 3D assets with additional geometric control.}
\label{fig:framework}
\end{figure}

\subsection{\shortname}
As shown in Figure~\ref{fig:framework}, \shortname is a unified framework for the controllable generation of 3D assets, which inherits the structure of Hunyuan3D 2.1~\cite{hunyuan3d2025hunyuan3d21imageshighfidelity}. In contrast, \shortname constructs a unified control encoder to introduce additional control signals, including point cloud, voxel, skeleton, and bounding box.

In this section, we begin by introducing the details of these controls, which significantly benefit in understanding the function of each additional control signal. Following this, we describe how to unify and interpolate these signals into the base model to achieve a unified controllable model.

\subsubsection{Skeleton Condition}

Unlike conventional object data, such as furniture, character data involves the definition of poses. In 3D animation production, modelers typically rig and skin characters in their rest pose to facilitate animation. On the other hand, in the design and printing of 3D figurines, we often create more dynamic and elaborate poses. Therefore, for an input image depicting an arbitrary pose, we may want to obtain a 3D model in a specified pose. To address this issue, we introduce the skeleton condition. We follow the PoseMaster~\cite{yan2025posemaster} approach to define the representation of the skeleton. Specifically, we use the 3D coordinates of the starting points of the bones to represent each bone, selecting both body bones and hand bones. Consequently, we can define the input for the skeleton condition as $P_{pose} \in \mathbb{R}^{M \times 6}$, where $M$ is the number of bones. We follow the PoseMaster's training strategy to construct the training pair, which randomly samples one frame to obtain the image and extracts the skeleton and mesh from the other.

\subsubsection{Bounding Box Condition}
In addition to specifying the pose of a character, defining the aspect ratio of the generated object in canonical space not only helps address the issue of overly thin geometry caused by the lack of thickness information in images, such as 2D cartoon, but also allows for geometric editing by adjusting the proportions, such as modifying the length, width, and height of a table. To achieve this, we propose the bounding box condition to control the aspect ratio. Specifically, we convert the length, width, and height ratios into the coordinates of the eight vertices in canonical space, which facilitates the design of a shared condition encoder. Consequently, we can define the input for the bounding box condition as $P_{box} \in \mathbb{R}^{8 \times 3}$. During training, we randomly perturb the rendered image or the underlying point cloud so that there is misalignment between the image and the point cloud, and then extract a bounding box from the point clouds as training input. 

\subsubsection{Point Cloud Condition}
The lighting, occlusion, and perspective of an image significantly influence the current 3D generation models. Under certain special viewpoints, even humans may struggle to accurately assess the geometry, making it a challenge to recover precise geometry from a single image. Therefore, it is essential to leverage additional input information to enhance the accuracy of the generated output, such as depth maps and LiDAR points. Unlike images, these two representations provide accurate spatial structural information, which aids the model in perceiving the spatial structure of the target model and consequently generating more accurate results. To this end, we introduce the point cloud condition.

Specifically, we consider that point clouds may originate from various sources, such as reconstruction models (e.g., VGGT~\cite{wang2025vggt}), LiDAR scans, and RGBD sensors. Typically, point clouds captured by LiDAR and RGBD cameras are incomplete. To simulate this type of data, we inherit a random drop sampling strategy from point cloud completion methods~\cite{yan2022fbnet,yu2021pointr,yan2025symmcompletion}. Additionally, we also support the input of complete point clouds, which can be obtained from reconstruction models. We directly use the 3D coordinates of the point clouds in space as their representation, employing three resolutions: 512, 1024, or 2048. To simulate noisy point clouds from depth sensors and LiDAR scans, we further apply noise perturbation by generating noise and adding it to the condition point cloud. Therefore, we can represent the point cloud condition as $P_c \in \mathbb{R}^{N_c \times 3}$, where $N_c$ is the number of points.

\subsubsection{Voxel Condition}
Similar to point clouds, voxels also serve as a strong conditioning signal, and we represent each voxel by the coordinates of its center. During Training, we construct a voxel condition from the surface point cloud. We uniformly sample from the object surface to obtain the initial point cloud $\text{P}$, and then convert the point cloud into voxels. First, we normalize the coordinates of $\text{P}$ to $\left[0, 16\right]^3$, quantize them into integers, and remove duplicate coordinates to obtain voxel coordinates at a resolution of $16\times16\times16$. Then, we map these coordinates back to the center coordinates of the voxels in the $[-1, 1]^3$ space.

\subsubsection{Unified Control Encoder}
The unified control encoder is used to unify and distinguish these extra control signals. As all control signals can be represented as a type of point cloud, we can design a shared point encoder to extract the corresponding features easily. In particular, we leverage the bone points as the skeleton representation to control the pose of the humanoid input. To align the dimensions of all extra conditions, we repeat the feature channel of the condition of voxel, point cloud, and bounding box. As a result, all controls can be represented as a point cloud $P_c \in \mathbb{R}^{N \times 6}$, where $N$ is the number of points or bones.

We then apply a position embedding followed by a linear layer on the point cloud $P_c$ to extract the condition features. Additionally, Different conditions correspond to distinct control objectives. For example, the point cloud condition aims to provide depth information from the image, reducing geometric distortions caused by artifacts, without altering the corresponding geometry of the image. In contrast, the skeleton condition is designed to control the pose of the input image, which does modify the corresponding geometry. Meanwhile, all these conditions are represented by point clouds, which tends to lead to control confusion. Therefore, to differentiate between the various conditions, we employ an embedding function to construct embeddings for the different control signals. Finally, we aggregate the extracted condition features and the distinguished embedding to obtain the final condition features. Thus, we can formulate our shared control encoder as follows:
\begin{align}
\beta_i = [\text{Linear}(\text{PosEmb}(P_i^c)), \mathcal{R}(\mathcal{M}(E(i)), r)], i \in [0,1,2,3]
\end{align}
where $\beta_i \in \mathbb{R}^{N \times C}$ is final control feature. $P_i^c$ is the control condition. ${E}$ and $M$ are the embedding function and the linear projection, respectively. $\mathcal{R}$ is a repeat operation, and $r$ is the time of repeating, which is used to enhance the signal for the condition type. $[\cdot]$ is a concatenation operation.

Benefiting from the scalability of the attention structure, we can concatenate the image feature $c_i$ and control feature $\beta$ to form a joint feature that is fed into the DiT model of Hunyuan3D 2.1. As a result, we can define the training objective of our \shortname as follows:
\begin{equation}
\mathbb{E}_{t, {x}_0, {x_1}, c'}\vert\vert{v}_\theta({x}, t, c')-(x_1-x_0)\vert\vert_2^2
\end{equation}
where $c'=[c, \beta_i]$ is the joint feature.

\section{Experiment}

\subsection{Implementation Details}

We follow the PoseMaster~\cite{yan2025posemaster} approach to construct the training data for pose control and adopt the dataset of Hunyuan3D 2.1~\cite{hunyuan3d2025hunyuan3d21imageshighfidelity} for other conditions.  We follow the setting of Hunyuan3D 2.1~\cite{hunyuan3d2025hunyuan3d21imageshighfidelity} to train our \shortname. Due to the varying lengths of different conditions, we set the batch size to 1. Additionally, we employ a random sampling strategy to select the control conditions for the current batch. Notably, since the data for the pose condition is less abundant and more challenging to learn, we use a higher sampling probability to prioritize tasks related to pose control. We directly train our diffusion model and unified control encoder using a fixed learning rate of 1e-5 and the AdamW optimizer. we utilize DINO-v2-Large~\cite{oquab2023dinov2} to extract image features.

\subsection{Qualitative Results}
In this section, we showcase the qualitative results for our \shortname. Specifically, we first use our \shortname to generate the controlled result and then leverage the Hunyuan3D 2.5~\cite{lai2025hunyuan3d2_5} to refine its geometry.

\begin{figure}[h]
    \centering
    \includegraphics[width=0.95\linewidth]{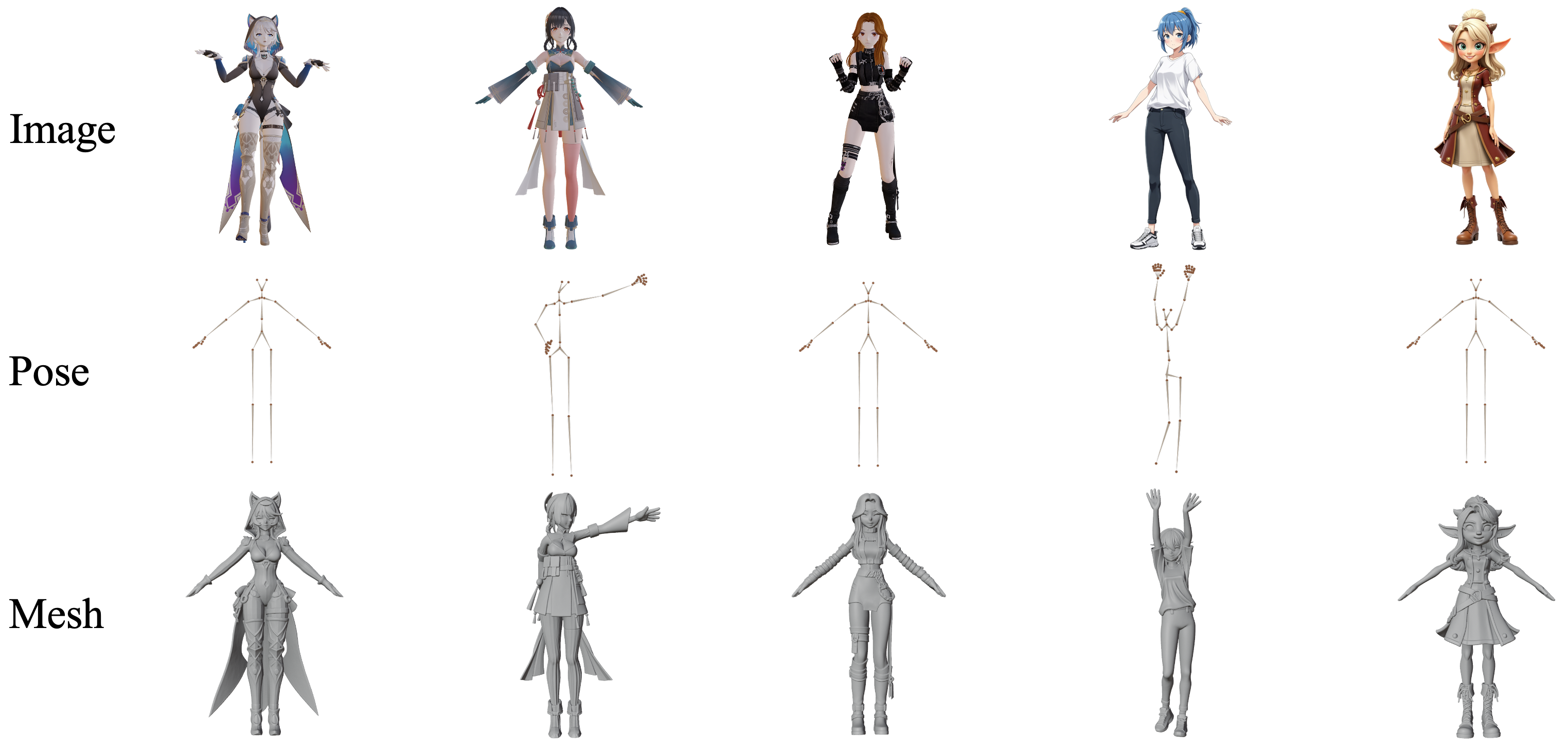}
    \caption{Qualitative results of pose control, demonstrating that our Omni model generates accurately aligned geometry across diverse poses and input styles.}
    \label{fig:exp_pose_control}
\end{figure}

\begin{figure}[h]
    \centering
    \includegraphics[width=0.95\linewidth]{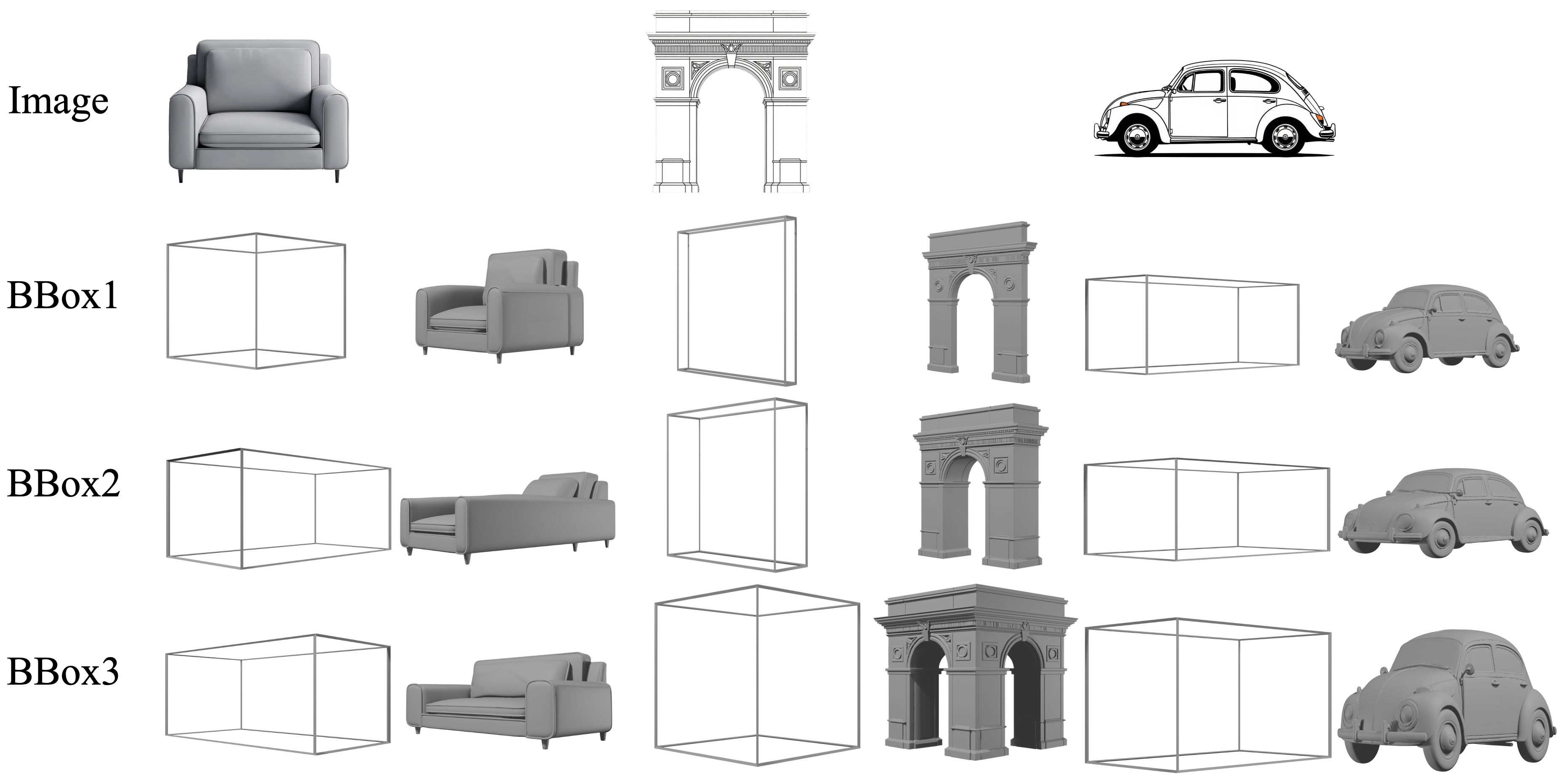}
    \caption{Qualitative results of bounding box control, our Omni model can generate objects at different scales, given different bounding boxes. }
    \label{fig:image-generation bbox ratio}

    \includegraphics[width=0.95\linewidth]{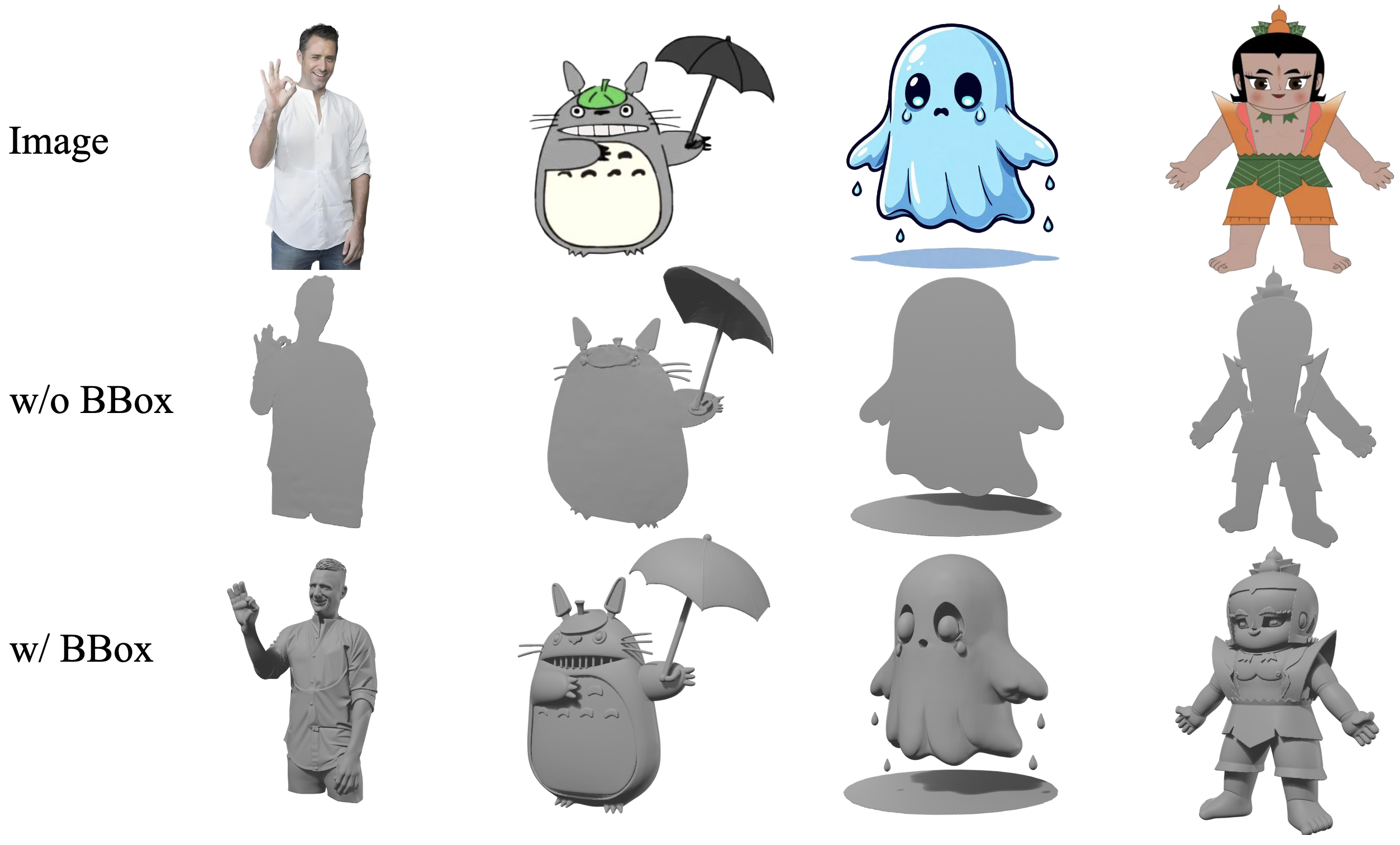}
    \caption{Qualitative results of bounding box control, given the bounding box condition, our Omni model can avoid generating thin, sheet-like geometry.}
    \label{fig:image-generation bbox compare}
\end{figure}

\begin{figure}[h]
    \centering
    \includegraphics[width=0.95\linewidth]{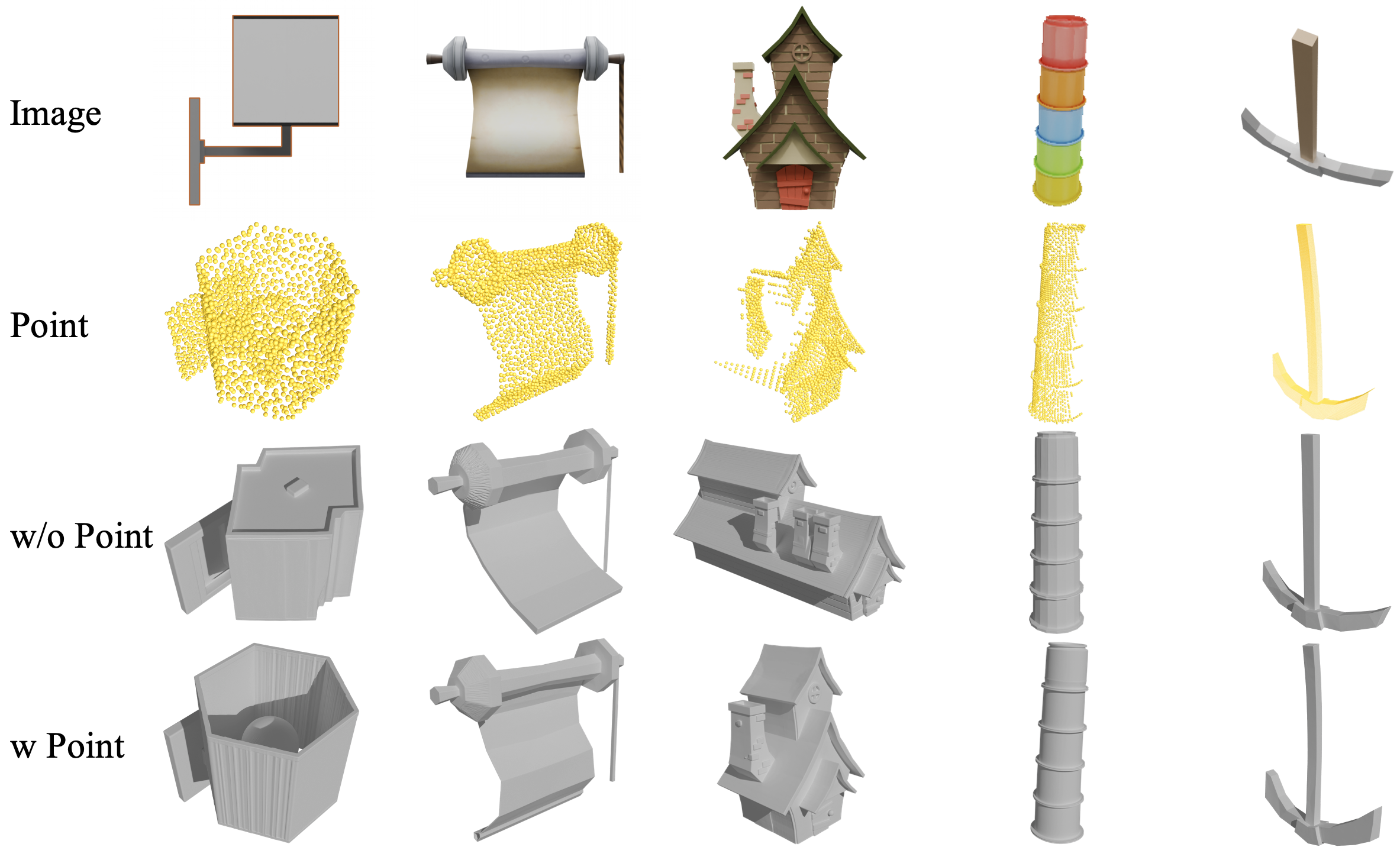}
    \caption{Qualitative results of point cloud control, showing improved alignment and detail recovery across complete, depth-projected, and scanned point cloud conditions compared to image-only results.}
    \label{fig:exp_point_control}
\end{figure}

\begin{figure}[h]
    \centering
    \includegraphics[width=0.95\linewidth]{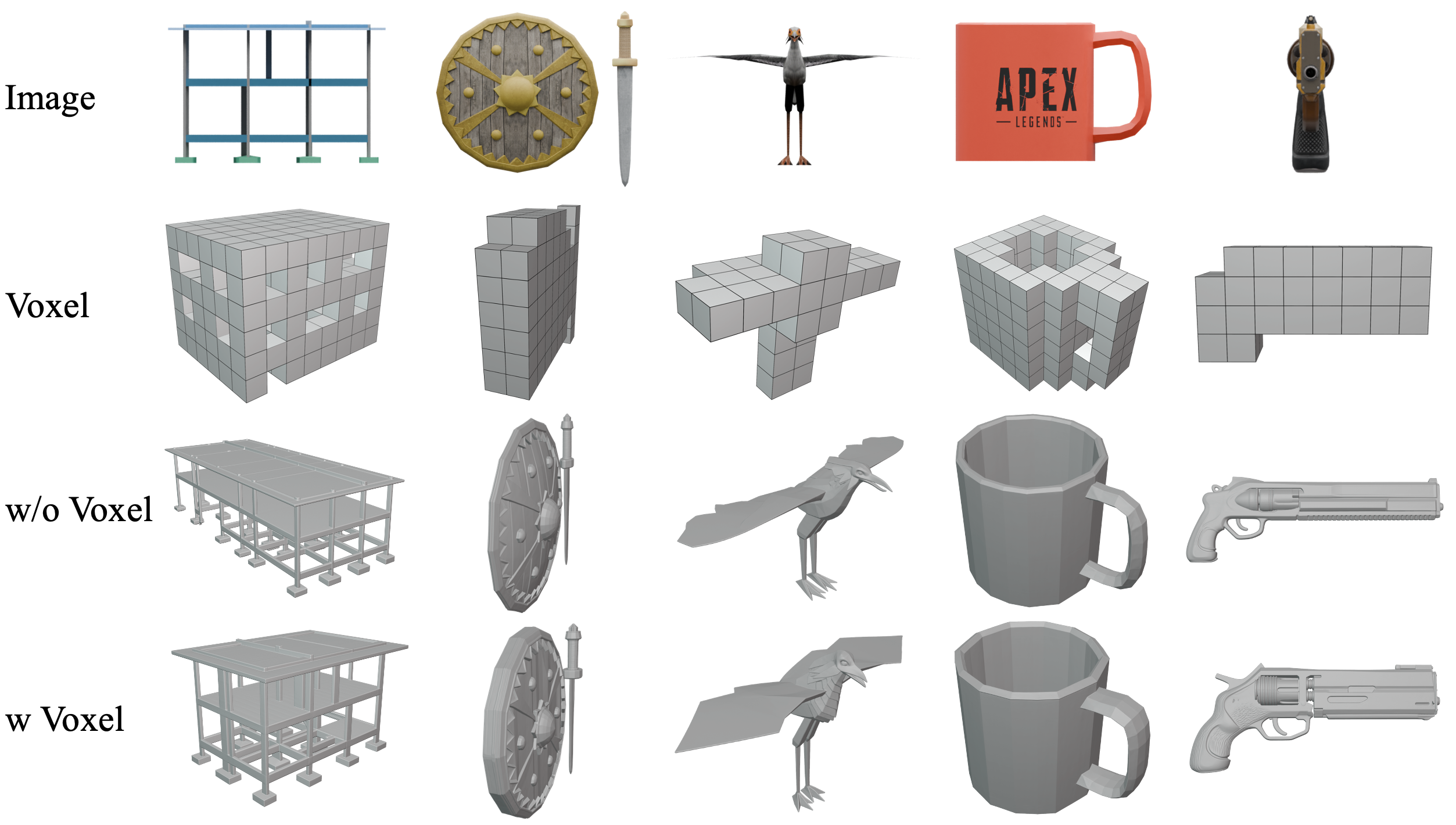}
    \caption{Qualitative results of voxel control, demonstrating improved scale alignment and fine detail recovery with voxel conditions compared to image-only results.}
    \label{fig:exp_voxel_control}
\end{figure}

{\bf Skeleton condition}. In contrast to the point cloud condition and voxel condition that provide sparse 3D information to enhance geometry accuracy, the skeleton condition aims to adjust the pose of the input image, which is usually used in character generation. As shown in Figure~\ref{fig:exp_pose_control}, with the skeleton condition as additional input, our Omni model can generate high-quality character geometry which accurately corresponds to the target pose, including A pose, sky pose, and hands-up pose. In our experiments, we use character images of various styles as conditional inputs, including rendered images from 3D character data and synthetic images produced by generative models. Regardless of the input style, our Omni model consistently produces human body meshes with fine geometric details that remain strictly aligned with the input skeleton, without artifacts. This strong capability for pose control demonstrates significant practical value for downstream applications such as 3D animation production and 3D figurine printing.

{\bf Bounding box condition}. Bounding box control signals allow flexible adjustment of the aspect ratio of the generated object. As shown in Figure~\ref{fig:image-generation bbox ratio}, given the same image condition, different bounding boxes successfully regulate the size of the output mesh. Notably, this manipulation is not a naive stretching: when the sofa is lengthened, extra supporting legs appear, and the Arc de Triomphe likewise acquires a plausible shape. Moreover, as shown in Figure~\ref{fig:image-generation bbox compare}, the bounding box signal can inject an activation cue into the generation network when single-image conditioned generation fails, yielding a valid mesh.

{\bf Point cloud condition}. 
As shown in Figure~\ref{fig:exp_point_control}, we present the generation results under two settings: image only and image with point cloud control. For the latter, we further consider three types of point cloud inputs: complete point clouds, point clouds from depth images, and point clouds from scans. For the first two cases in the figure, we observe that providing a complete point cloud as a control signal effectively resolves the ambiguity inherent in single-view inputs and allows the recovery of occluded internal structures. In the third and fifth cases, where surface point clouds are obtained via a depth map, the additional input similarly mitigates single-view ambiguity, ensuring that the generated geometry is well-aligned in scale with the ground truth. In the fourth case, given a noisy surface point cloud from a scan, the generated geometry is also better aligned with the original object compared with the image-only baseline, addressing the issue where the image encoder tends to ignore the true object pose. In summary, once point cloud input is provided, our Omni model can effectively align the generated geometry with real-world geometry. This further demonstrates that even partial point clouds serve as a strong cue for improving the quality of 3D geometry generation.

{\bf Voxel Condition}. 
Similar to the point cloud condition, the voxel condition provides sparse geometric cues that help resolve the ambiguities inherent in a single image. As shown in Figure~\ref{fig:exp_voxel_control}, in the first and fifth cases, the additional voxel control condition ensures that the generated objects are properly aligned in scale with the ground truth geometry. Cases 2, 3, and 4 further illustrate how the voxel condition contributes to recovering fine geometric details. For instance, restoring the flat surface of the shield, capturing the shape of the bird's wing, and reproducing the low-poly style geometry of the cup. These examples clearly demonstrate that incorporating voxel conditions enables the model to faithfully recover both the proportions and the details of object geometry, thereby further improving generation quality.

\section{Conclusion}
In this paper, we propose a unified framework, called \shortname, for fine-grained and controllable 3D asset generation. We incorporate point cloud, voxel, bounding box, and skeleton to mitigate geometry distortion in image-only 3D generation and achieve style control. To unify these extra conditions in one diffusion model, we design a lightweight unified control encoder. Building on the powerful existing 3D generation model, such as Hunyuan3D 2.1, \shortname just introduces a lightweight encoder to achieve high-quality controllable generation. Experiments show that these additional controls improve generation accuracy, enable geometry-aware transformations, and increase robustness for production workflows.

\section{Contributors}
\large{Authors are listed \textbf{alphabetically by the first name}.} 
\definecolor{tencentblue}{RGB}{38,54,221}

\large{
\color{tencentblue}%
\begin{multicols}{2}
\raggedcolumns
Bowen Zhang\\
Chunchao Guo\\
Haolin Liu\\
Hongyu Yan\\
Huiwen Shi\\
Jingwei Huang\\
Junlin Yu\\
Kunhong Li\\
Linus\\
Penghao Wang\\
Qingxiang Lin\\
Sicong Liu\\
Xianghui Yang\\
Yixuan Tang\\
Yunfei Zhao\\
Zeqiang Lai\\
Zhihao Liang\\
Zibo Zhao
\end{multicols}}

\clearpage

\bibliography{colm2024_conference}
\bibliographystyle{colm2024_conference}

\end{document}